\documentclass[10pt,twocolumn,letterpaper]{article}

 \usepackage[pagenumbers]{wacv} % To force page numbers, e.g. for an arXiv version

\usepackage{graphicx}
\usepackage{amsmath}
\usepackage{amssymb}
\usepackage{booktabs}
\usepackage{algorithm}
\usepackage{algorithmic}
\usepackage{multirow}
\usepackage{colortbl}
\usepackage{float}
\usepackage[accsupp]{axessibility} % Improves PDF readability for those with disabilities.
\usepackage[pagebackref,breaklinks,colorlinks]{hyperref}

\usepackage[capitalize]{cleveref}
\crefname{section}{Sec.}{Secs.}
\Crefname{section}{Section}{Sections}
\Crefname{table}{Table}{Tables}
\crefname{table}{Tab.}{Tabs.}

\def\wacvPaperID{264} % *** Enter the WACV Paper ID here
\def\confName{WACV}
\def\confYear{2025}

\definecolor{firstcolor}{RGB}{252, 235, 193}%{236 168 169}
\definecolor{secondcolor}{RGB}{241, 233, 223} %{211 226 183}
\definecolor{thirdcolor}{RGB}{219, 200, 189} %{116 174 212}
\definecolor{thirdcolor}{RGB}{255, 255, 255}

\newcommand{\markfirst}[1]{\cellcolor{firstcolor}{\textbf{#1}}}
\newcommand{\marksecond}[1]{\cellcolor{secondcolor}{#1}}

\newcommand{\colorfirsttext}[1]{\colorbox{firstcolor}{\textbf{#1}}}
\newcommand{\colorsecondtext}[1]{\colorbox{secondcolor}{#1}}

\begin{document}

% \title{OmniGS: Omnidirectional Gaussian Splatting for Fast Radiance Field Reconstruction using Omnidirectional Images}
\title{OmniGS: Fast Radiance Field Reconstruction using Omnidirectional Gaussian Splatting}

\author{First Author\\
Institution1\\
Institution1 address\\
{\tt\small firstauthor@i1.org}
\and
Second Author\\
Institution2\\
First line of institution2 address\\
{\tt\small secondauthor@i2.org}
}

\author{Longwei Li$^1$ \quad Huajian Huang$^2$ \quad Sai-Kit Yeung$^2$ \quad Hui Cheng$^1$\\
$^1$Sun Yat-sen University \quad
$^2$ The Hong Kong University of Science and Technology\\
{\tt\small lilw23@mail2.sysu.edu.cn},
{\tt\small hhuangbg@connect.ust.hk},
{\tt\small saikit@ust.hk},
{\tt\small chengh9@mail.sysu.edu.cn}\\
}

\twocolumn[{
\maketitle
\begin{figure}[H]
% \begin{center}
    \hsize=\textwidth
    \centering
    \vspace{-20pt}
    \includegraphics[width=1.99\linewidth]{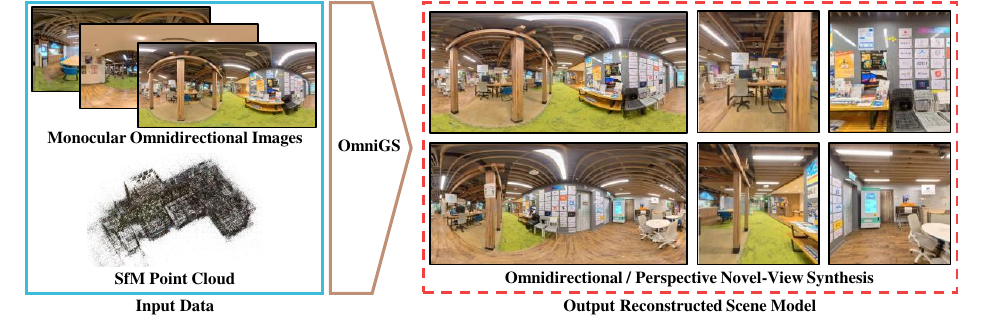}
    % \vspace{-5pt}
    \caption{We introduce OmniGS, a novel omnidirectional radiance field reconstruction method. It takes a series of calibrated monocular 360-degree images and sparse SfM point clouds as input to quickly recover 3D Gaussians as large-scale scene representations, achieving real-time omnidirectional novel view synthesis.
    %OmniGS is a novel omnidirectional  reconstruction method. It takes a series of calibrated monocular 360 images and their sparse SfM point cloud as input, and reconstructs the scene using 3D Gaussians as representations, achieving real-time novel-view synthesis.
    }\label{fig:headfigure}
\end{figure}
% \end{center}
}]

% \setlength{\textfloatsep}{8pt}

%%%%%%%%% ABSTRACT
\begin{abstract}
% \vspace{-0.1cm}
Photorealistic reconstruction relying on 3D Gaussian Splatting has shown promising potential in various domains. However, the current 3D Gaussian Splatting system only supports radiance field reconstruction using undistorted perspective images. In this paper, we present OmniGS, a novel omnidirectional Gaussian splatting system, to take advantage of omnidirectional images for fast radiance field reconstruction. Specifically, we conduct a theoretical analysis of spherical camera model derivatives in 3D Gaussian Splatting. According to the derivatives, we then implement a new GPU-accelerated omnidirectional rasterizer that directly splats 3D Gaussians onto the equirectangular screen space for omnidirectional image rendering. We realize differentiable optimization of the omnidirectional radiance field without the requirement of cube-map rectification or tangent-plane approximation. Extensive experiments conducted in egocentric and roaming scenarios demonstrate that our method achieves state-of-the-art reconstruction quality and high rendering speed using omnidirectional images. The code will be publicly available at \url{https://github.com/liquorleaf/OmniGS}. %To benefit the research community, t once the paper is published.
\end{abstract}

%%%%%%%%% BODY TEXT
\section{Introduction}
\label{sec:intro}
%photo realistic reconstruction -> 3D splatting -> how to do, or what is 3D gaussian splatting -> application, e.g., -> but only for undistorted pinhole image, why need 360 splatting/the advantage of 360-degree image -> concurrent work 360GS limitation -> our solution

Reconstructing three-dimensional (3D) structures of observed environments plays an important role in many applications, such as environmental monitoring, virtual reality, localization, navigation, path planning, and other high-level perception tasks. Recent progress \cite{review:360recons} in this realm has sought to harness the information contained in large field-of-view (FoV) images for efficient reconstruction, especially omnidirectional images which are able to capture the entire environment at each shot. 
%Owing to a 360-degree FoV, an omnidirectional camera is able to capture the entire environment at each shot, providing consistent observations. Some SLAM systems such as 360VO \cite{sparse:360vo}, 360VIO \cite{sparse:360vio}, and LF-VISLAM \cite{sparse:lfvislam} make use of consistent observations to achieve robust pose estimation and efficient reconstruction. 
%However, the conventional use of point clouds or meshes to model reconstructed environments, although effective in capturing structural information, often fails to provide appealing visualization. %visually

In order to achieve photorealistic reconstruction and enable immersive scene roaming using omnidirectional images, several approaches, such as 360Roam~\cite{nerf:360roam}, 360FusionNerf~\cite{ nerf:360fusionnerf}, and PaniGRF~\cite{nerf:panogrf} have explored the utilization of the neural radiance field (NeRF) technique \cite{nerf}. 
Unlike traditional methods~\cite{sparse:360vo, sparse:360vio, sparse:lfvislam} that focus on precise geometry reconstruction, NeRF-based methods employ multi-layer perceptrons (MLPs) to implicitly model the scene. It accumulates density and view-dependent color per ray, which are regressed from the MLPs, for image synthesis. Meanwhile, the MLP parameters are optimized by minimizing the photorealistic loss between rendered images and corresponding training images. Since such a rendering and optimization process requires millions of ray samplings, NeRF-based methods~\cite{nerf:360roam, nerf:360fusionnerf, nerf:panogrf} suffer from long training or inference time to model omnidirectional radiance field.

Recently, 3D Gaussian Splatting (3DGS) \cite{3dgs} effectively addresses the limitation of NeRF by introducing 3D Gaussians to explicitly represent radiance field.
Each 3D Gaussian is a point associated with certain attributes, i.e. position, color, scale, rotation, and opacity. For rendering, the {elliptical weighted average (EWA)} splatting algorithm \cite{3dgs:ewasplatting} is applied to project and rasterize 3D Gaussians onto the screen space. Benefiting from the highly efficient rendering process, 3DGS significantly reduces training and inference time in high-quality radiance field reconstruction, having great potential in real-time applications~\cite{3dgs:splatam,3dgs:gsslam,3dgs:gaussianslam,3dgs:photoslam}. Nevertheless, the current splatting algorithm is only compatible with undistorted perspective image rendering. 

%required for accurate representation and  of.
In this paper, we aim to propose a novel system to tame 3D Gaussian splatting for fast omnidirectional radiance field reconstruction. 
To achieve differentiable omnidirectional image rendering, we begin by conducting a theoretical analysis of the derivatives of the spherical camera model in 3D Gaussian splatting. 
Building upon the derived derivatives, we develop a new GPU-accelerated omnidirectional rasterizer that directly splats 3D Gaussians onto the equirectangular screen space without the need for cube-map rectification or tangent-plane approximation. The omnidirectional rasterizer then builds up the foundation of our fast omnidirectional radiance field reconstruction system, named OmniGS. OmniGS efficiently recovers the radiance field from omnidirectional images for novel view synthesis, as shown in \cref{fig:headfigure}. 
%We provide the mathematical foundation forming the foundation of 
To verify the efficacy of our proposed system, we conducted extensive evaluations on the omnidirectional roaming scenes of 360Roam \cite{nerf:360roam} and egocentric scenes of EgoNeRF \cite{nerf:egonerf}. 
Qualitative and quantitative results show that our method achieves state-of-the-art (SOTA) performance regarding photorealistic reconstruction quality and rendering speed using omnidirectional images. 

In summary, our contributions are as follows:
\begin{itemize}
    \item We introduce thoughtful theoretical analysis of the omnidirectional Gaussian Splatting, enabling direct splatting of the 3D Gaussians onto the equirectangular screen space for real-time and differentiable rendering.
    %\item We build a novel omnidirectional photorealistic 3D reconstruction system, named OmniGS, rooted in mathematical analysis of the derivatives and gradients of our rendering method. The system achieves fast and high-fidelity photorealistic reconstruction.
    \item We develop OmniGS, a novel photorealistic reconstruction system based on our new GPU-accelerated omnidirectional rasterizer. %achieving fast and high-fidelity reconstruction.
    %enabling real-time forward rendering and rapid loss backpropagation. 
    \item The extensive experiments demonstrate that our system achieves state-of-the-art omnidirectional radiance field reconstruction quality and fast rendering speed. 
    %The code of OmniGS will be publicly available.
\end{itemize}

%------------------------------------------------------------------------
\section{Related Works}
\label{sec:related-works}
%Recent advancements have focused on 
\subsection{Omnidirectional Reconstruction}
Reconstructing 3D structures of observed environments is a fundamental task that often relies on multi-view geometry and factor graph solvers. Leveraging the information embedded in large field-of-view (FoV) images can facilitate efficient and robust reconstruction. Existing methods such as OpenMVG \cite{sparse:openmvg} utilize feature points on spherical images to establish 2D-to-3D correspondences and optimize the 3D structure by minimizing spherical reprojection errors. 
Additionally, techniques such as 360VIO \cite{sparse:360vio} and LF-VISLAM \cite{sparse:lfvislam} integrate inertial measurement units (IMUs) into the visual system, leading to improved accuracy and robustness in estimating motion structures. In contrast, 360VO \cite{sparse:360vo} pioneers direct visual odometry using a monocular omnidirectional camera, producing semi-dense reconstructed point cloud maps. \cite{Egocentric:SIG:2022} proposes an egocentric 3D reconstruction method that can acquire scene geometry with high accuracy from a short egocentric omnidirectional video. 
%However, these methods often yield sparse maps that fall short in realistic exploration scenarios. 
However, the conventional use of point clouds or meshes to model reconstructed environments, although effective in capturing structural information, often fails to provide appealing visualization.

\subsection{Omnidirectional NeRF Reconstruction}
%PanoGRF: Generalizable Spherical Radiance Fields for Wide-baseline Panoramas
%PERF: Panoramic Neural Radiance Field from a Single Panorama utilizes multiple 360-degree images to progressively reconstruct explicit geometry. It further
%
With the advancements in neural radiance field (NeRF) techniques, photorealistic reconstruction has gained increased flexibility. 
360Roam \cite{nerf:360roam} first introduced the use of omnidirectional radiance fields for immersive scene exploration.
To improve NeRF performance in large-scale scenes, 360Roam incorporates geometry-aware sampling and decomposition of the global radiance field, resulting in fast and high-fidelity synthesis of novel views. 360FusionNeRF \cite{nerf:360fusionnerf} and PERF \cite{nerf:perf} aim to reconstruct the radiance field from a single omnidirectional image, alleviating the need for a large training dataset. However, due to the limited information provided by the single image, additional depth maps are necessary to enhance the performance of novel view synthesis, particularly in complex scenes.
To address the issue of training view overfitting in NeRF, PanoGRF~\cite{nerf:panogrf} integrates deep features and 360-degree scene priors into the omnidirectional radiance field. Contrasting the roaming scenarios explored by previous methods, EgoNeRF \cite{nerf:egonerf} focuses on egocentric scenes captured within a small circular area using casually acquired omnidirectional images. EgoNeRF employs quasi-uniform angular grids to adaptively model unbounded scenes, achieving state-of-the-art performance in egocentric view synthesis.
Despite the success of NeRF-based methods in novel view synthesis using omnidirectional images, the computational intensity of radiance field sampling remains a challenge, leading to slow training or inference speeds. 

\subsection{3D Gaussian Reconstruction}
%Recently, 3D Gaussian Splatting \cite{3dgs:gsslam} demonstrates high training and rendering speed while achieving impressive rendering results.   with 3D Gaussian Splatting
Recently, the emergence of 3DGS \cite{3dgs} has opened up new possibilities for real-time applications. Building upon this foundation, GS-SLAM \cite{3dgs:gsslam} and Splatam \cite{3dgs:splatam} utilize 3D Gaussians as a scene representation to perform dense RGB-D SLAM. Similarly, Gaussian Splatting SLAM \cite{3dgs:monoGS} adopts direct optimization against 3D Gaussians for camera tracking. In addition, Photo-SLAM \cite{3dgs:photoslam} achieves top photorealistic mapping quality without relying on dense depth optimization. Notably, it can execute online mapping seamlessly on an embedding device at a real-time speed, highlighting its potential for robotics applications.

To extend Gaussian Splatting for omnidirectional image rendering, a concurrent work 360-GS \cite{3dgs:360gs} leverages a tangent-plane approximation to formulate the Gaussian splatting process. However, even though 360-GS outperforms previous NeRF-based baselines in terms of omnidirectional image rendering quality and training speed, its two-stage projection splatting method is suboptimal. Furthermore, its reliance on indoor layout priors limits its generalization capabilities in multi-room scale and outdoor scenarios.
In this paper, our proposed OmniGS uses direct screen-space splatting to accelerate rendering and does not rely on scene assumptions or deep networks, enabling its application in diverse indoor and outdoor scenes. 

%------------------------------------------------------------------------
\section{Omnidirectional Gaussian Splatting}

% In this section, we will introduce our omnidirectional Gaussian Splatting method, including the omnidirectional camera model, the forward rendering process, and the corresponding backward gradient propagation.

\subsection{Preliminary}\label{subsec:splatting-preliminary}
3D Gaussian Splatting\cite{3dgs} is an anisotropic point-based scene representation. Each point is parameterized by 3D center position $\mathbf{m}$, color $c$ (derived from $\mathbf{m}$ and Spherical Harmonics coefficients), rotation $\mathbf{q}$, scale $\mathbf{S}$ and opacity $o$, where $\mathbf{q}$ and $\mathbf{S}$ are used to derive the 3D Gaussian covariance of that point. To render an image from a certain view, all points and their covariance are projected, sampled, and $\alpha$-blended onto the 2D screen space. This forward rendering procedure is differentiable, allowing scene reconstruction and optimization from pose-calibrated images and a sparse point cloud.

\subsection{Camera Model}\label{subsec:splatting-camera-model}

\begin{figure}[t]
    \centering
        \subfloat[Camera coordinate system.]{
            \includegraphics[width=0.45\linewidth]{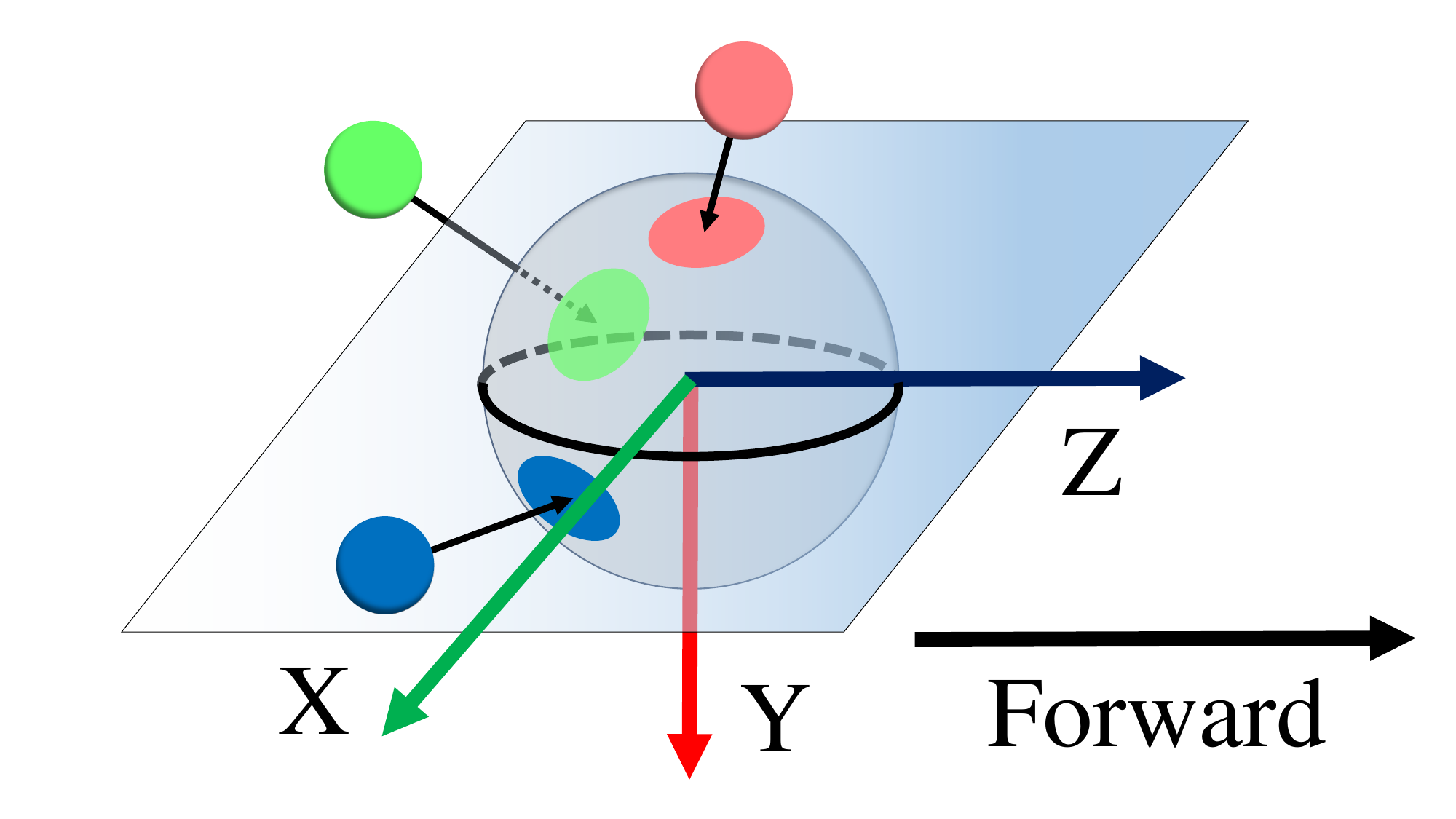}
            \label{fig:camera-camera}}
        \hfill
        \subfloat[Image coordinate system.]{
            \includegraphics[width=0.45\linewidth]{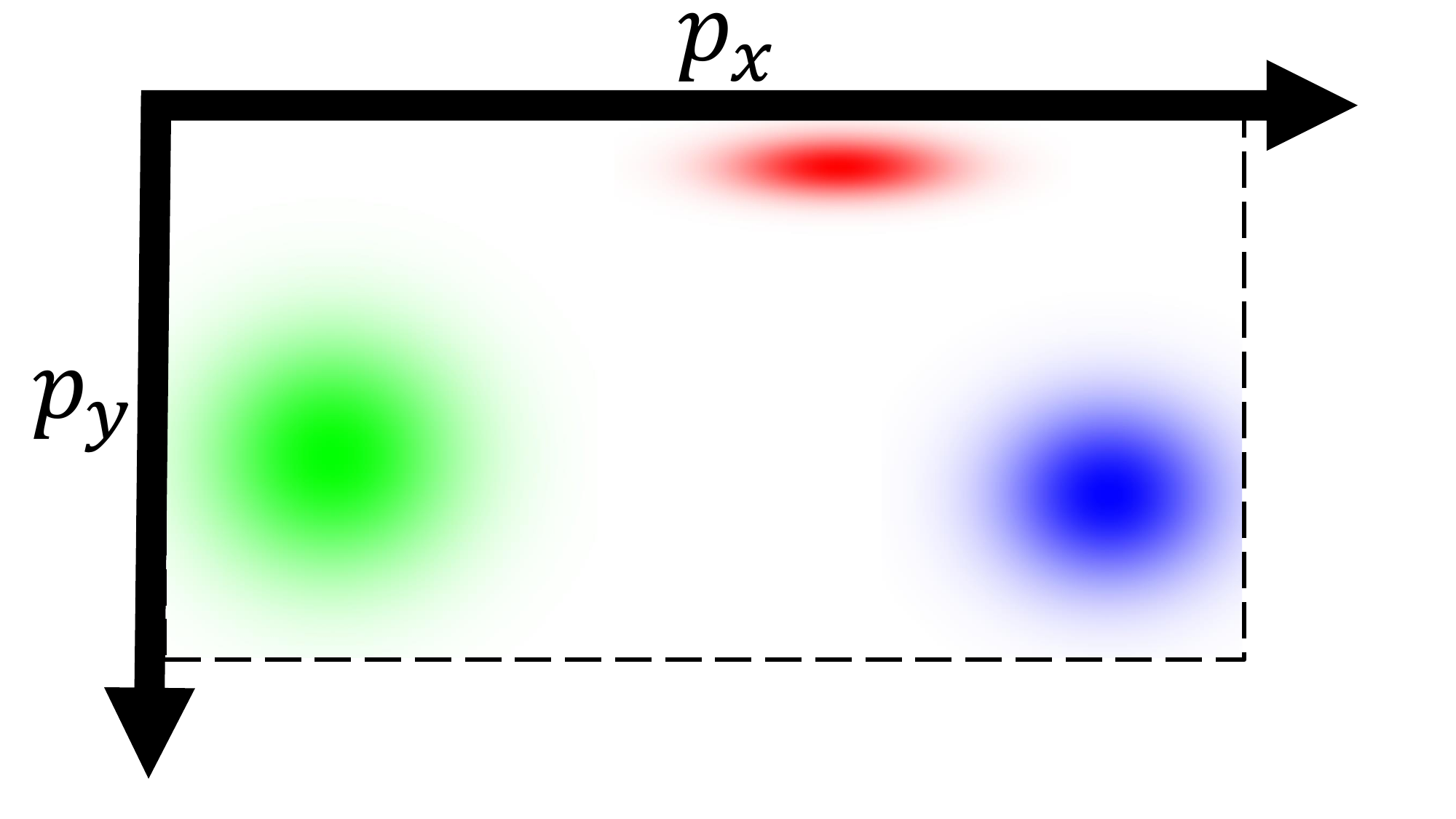}
            \label{fig:camera-pixel}}
        \\
        \subfloat[Latitude and longitude.]{
            \includegraphics[width=0.45\linewidth]{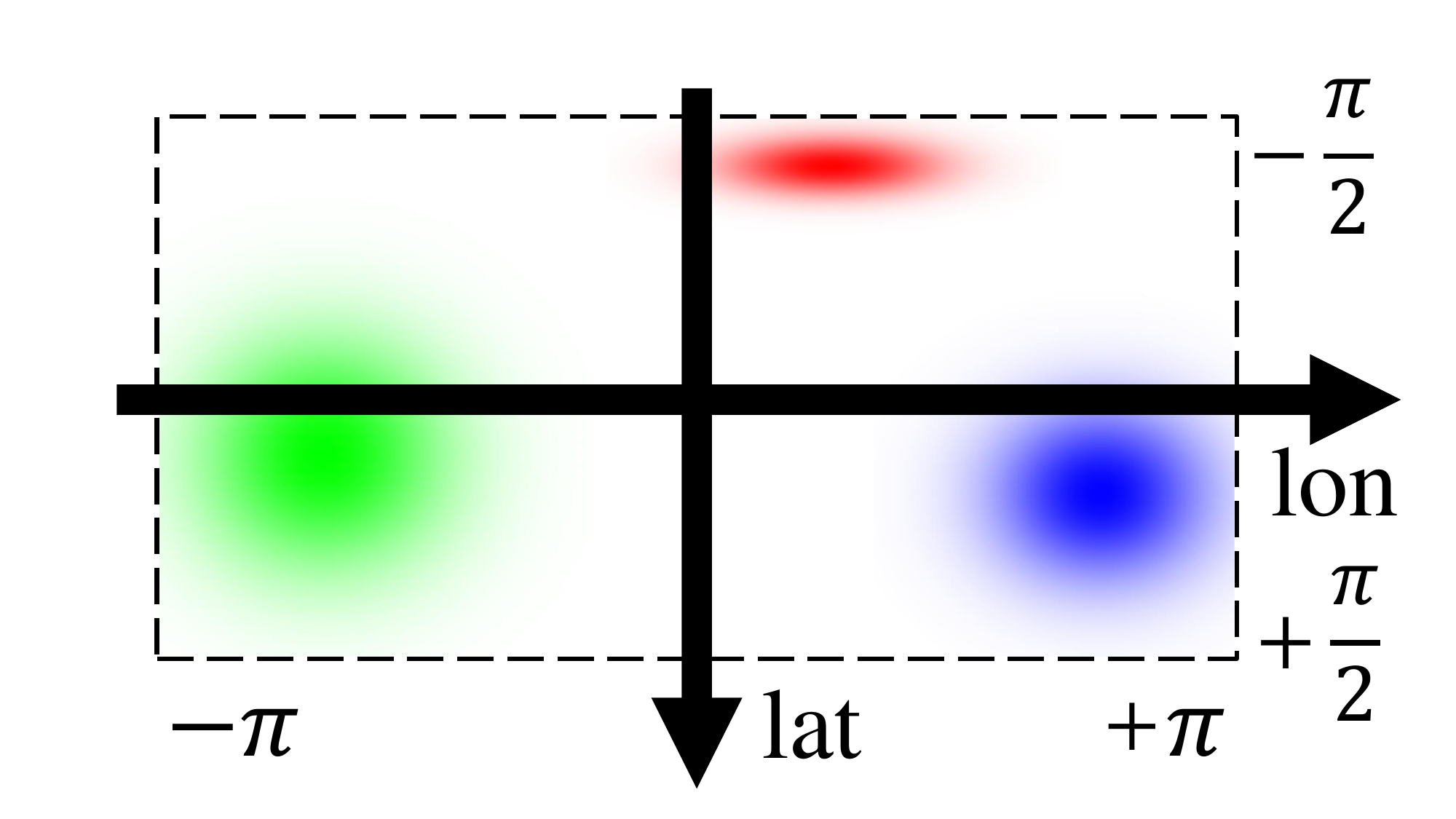}
            \label{fig:camera-lonlat}}
        \hfill
        \subfloat[Uniform screen-space.]{
            \includegraphics[width=0.45\linewidth]{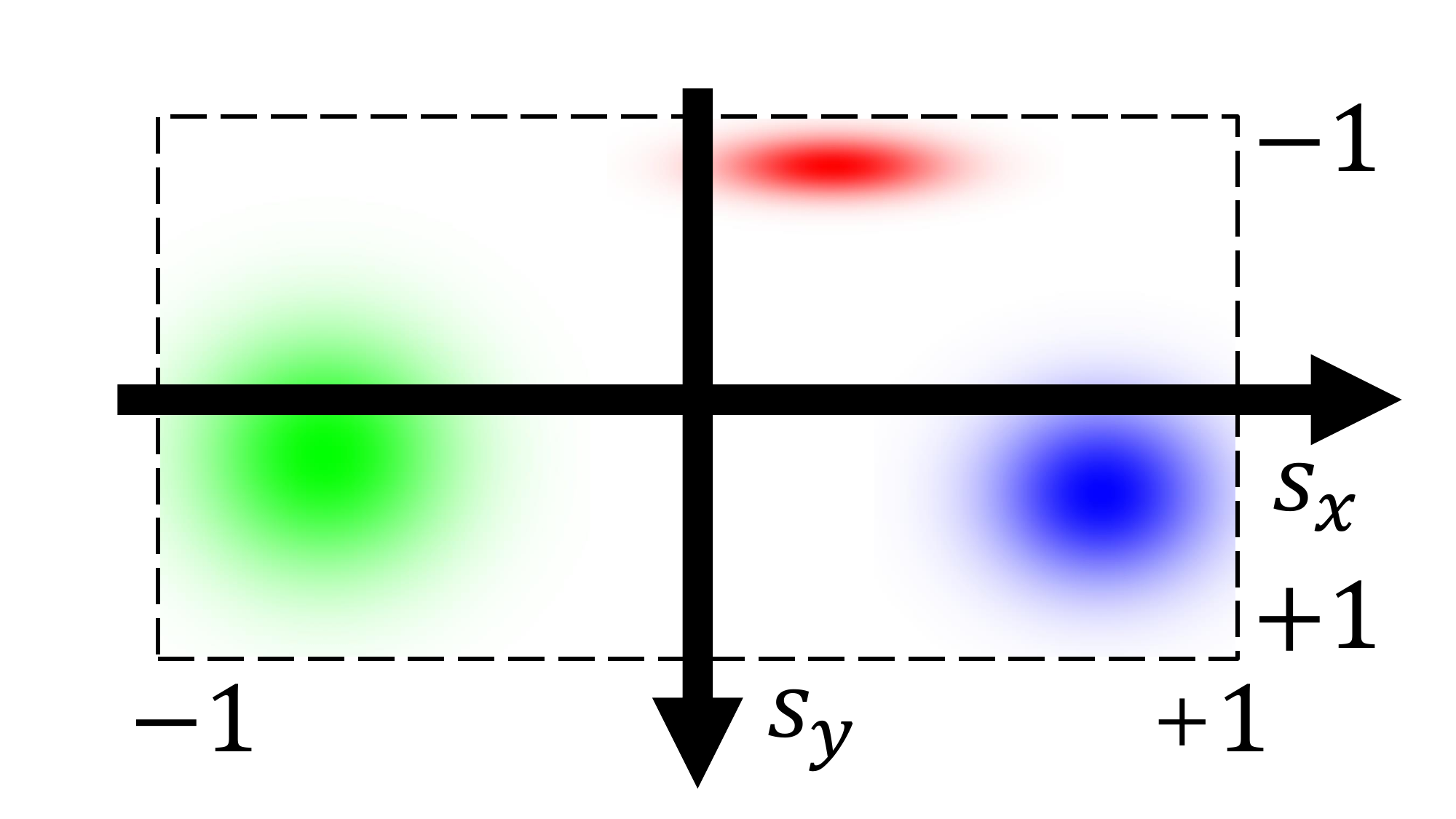}
            \label{fig:camera-screen}}
    \caption{Coordinate systems used in OmniGS. We use the SLAM convention for cameras, i.e. +X is right, +Y is down, and +Z is forward. In the forward rendering process, 3D Gaussians are first transformed from the world coordinate system to the camera space, then projected onto the image. The latitude-longitude coordinate system and the uniform screen-space coordinate system serve as intermediate variables during the projection process. The X-Z plane of the camera space is the equatorial plane, i.e. $lat=0$.}
    \label{fig:camera-model}
\end{figure}

The camera model is the mathematical relationship between a 3D camera-space point $\mathbf{t}=[t_x,t_y,t_z]^\text{T}$ and its projected position $\mathbf{p}=[p_x,p_y]^\text{T}$ on the image. Let $\mathbf{m}$ be the world position of the point, and $\mathbf{T}_\text{cw}$ be its transform from the world coordinate system to the camera space, we have
\begin{equation}
    \mathbf{t} = \mathbf{T}_\text{cw}*\mathbf{m} = \mathbf{W}\mathbf{m}+\mathbf{t}_\text{cw}.
\end{equation}
where $\mathbf{W}$ is the rotation matrix and $\mathbf{t}_\text{cw}$ is the translation.

The original 3DGS uses the perspective camera model:
\begin{equation}\label{equ:perspective-camera-model}
    \begin{bmatrix}
        p_x\\
        p_y
    \end{bmatrix}
    =
    \begin{bmatrix}
        f_x t_x/t_z + c_x\\
        f_y t_y/t_z + c_y
    \end{bmatrix},
\end{equation}
where $f_x,f_y$ are focal lengths and $c_x,c_y$ are the principle points of the pinhole camera model.

\begin{figure*}[t]
    \centering
    \includegraphics[width=\linewidth]{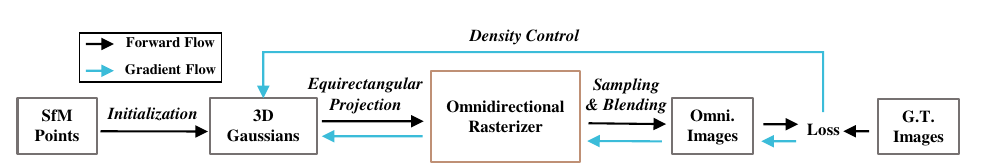}
    \caption{A schematic overview of OmniGS optimization flow. It optimized 3D Gaussian representation by minimicing the loss between the rendering omnidirectional images and the input ground truth images.}
    \label{fig:pipeline}
    % \vspace{-8pt}
\end{figure*}

To take advantage of one-shot omnidirectional images, we use the equirectangular projection model, which is the most commonly used form in the context of omnidirectional reconstruction. As shown in \cref{fig:camera-camera}, we define the camera coordinate system according to the SLAM convention. The camera X-Z plane corresponds to the equatorial plane of equirectangular projection. To keep high fidelity, we use the original inverse trigonometric functions to compute the spherical latitude $lat$ and longitude $lon$:
\begin{equation}\label{equ:camera-to-lonlat}
    \begin{bmatrix}
        lon\\
        lat
    \end{bmatrix}
    =
    \begin{bmatrix}
        \mathrm{arctan2}(t_x / t_z)\\
        \arcsin (t_y / t_r)
    \end{bmatrix},
\end{equation}
where $t_r = \sqrt{t_x^2 + t_y^2 + t_z^2}$ is the distance from the center of unit sphere to the center of the 3D Gaussian in the camera space, $\mathrm{arctan2}$ is the 4-quadrant inverse tangent, and we have $-\pi \leq lon < \pi$ and $-\pi/2 \leq lat < \pi/2$. Then the above latitude and longitude (\cref{fig:camera-lonlat}) can be transformed into the uniform screen-space coordinates (\cref{fig:camera-screen}):
\begin{equation}\label{equ:lonlat-to-screen}
    \begin{bmatrix}
        s_x\\
        s_y
    \end{bmatrix}
    =
    \begin{bmatrix}
        lon / \pi\\
        2lat / \pi
    \end{bmatrix},
\end{equation}
so that we have $-1 \leq s_x,s_y < 1$. At the end of the projection process, the uniform screen-space coordinates are transformed into the pixel position on the image (\cref{fig:camera-pixel}):
\begin{equation}\label{equ:screen-to-pixel}
    \begin{bmatrix}
        p_x\\
        p_y
    \end{bmatrix}
    =
    \begin{bmatrix}
        (s_x + 1) W / 2\\
        (s_y + 1) H / 2
    \end{bmatrix},
\end{equation}
where $W$,$H$ are the width and height of the equirectangular image respectively, measured in the count of pixels.

\subsection{Forward Rendering}\label{subsec:splatting-forward}
Following 3DGS~\cite{3dgs}, the final color of each image pixel is decided following the $\alpha$-blending model:
\begin{equation}\label{equ:alpha-blending}
    C=\sum_{i=1}^N c_i \alpha_i \prod_{j=1}^{i-1}(1-\alpha_j),
\end{equation}
where $N$ is the number of 3D Gaussians near this pixel. For perspective cameras, these Gaussians are sorted by their $t_z$, from nearest to farthest. However, under the circumstances of omnidirectional vision, the criterion for sorting is changed to the distance between the camera center and 3D Gaussian kernal center $t_r$. The $i$-th Gaussian has color $c_i$ and sampled intensity $\alpha_i$. Furthermore, $\alpha_i$ is determined by its opacity $o_i$ and the sampled value on its 2D Gaussian distribution:
\begin{equation}\label{equ:alpha-alpha}
    \alpha_i = o_i G_i(\Delta \mathbf{p}_i),
\end{equation}
where $\Delta \mathbf{p}_i = \mathbf{p}_i - \mathbf{p}_s$ is the difference vector between its projected center $\mathbf{p}_i$ and the sampling pixel position $\mathbf{p}_s$, and the sampling on the 2D Gaussian function is defined as:
\begin{equation}\label{equ:alpha-G}
    G_i(\Delta \mathbf{p}_i) = \exp\left({-\frac{1}{2} (\Delta \mathbf{p}_i)^\text{T} \Tilde{\mathbf{\Sigma}}^{-1}_i (\Delta \mathbf{p}_i)}\right).
\end{equation}

To get 2D covariance $\Tilde{\mathbf{\Sigma}}$ of the Gaussian projected onto the equirectangular image plane, we compute it according to the local affine approximation method described in \cite{3dgs:ewasplatting}:
\begin{equation}\label{equ:cov2D}
    \Tilde{\mathbf{\Sigma}} \approx \mathbf{J} \mathbf{W} \mathbf{\Sigma} \mathbf{W}^\text{T} \mathbf{J}^\text{T},
\end{equation}
where $\mathbf{\Sigma}$ is the 3D covariance, derived from the scaling vector and rotation quaternion of this Gaussian \cite{3dgs}, $\mathbf{J}$ is the Jacobian of the camera projection described in \cref{subsec:splatting-camera-model}:
\begin{equation}\label{equ:J}
    \mathbf{J} = 
    \begin{bmatrix}
        \dfrac{\partial p_x}{\partial t_x} & \dfrac{\partial p_x}{\partial t_y} & \dfrac{\partial p_x}{\partial t_z} \\[16pt]
        \dfrac{\partial p_y}{\partial t_x} & \dfrac{\partial p_y}{\partial t_y} & \dfrac{\partial p_y}{\partial t_y} \\[16pt]
        0 & 0 & 0
    \end{bmatrix}
    ,
\end{equation}
with
\begin{align}
    \frac{\partial p_x}{\partial t_x} &= +\frac{W}{2\pi}\cdot\frac{t_z}{{t_x^2 + t_z^2}}, \\
    \frac{\partial p_x}{\partial t_y} &= 0, \\
    \frac{\partial p_x}{\partial t_z} &= -\frac{W}{2\pi}\cdot\frac{t_x}{{t_x^2 + t_z^2}}, \\
    \frac{\partial p_y}{\partial t_x} &= -\frac{H}{\pi}\cdot\frac{t_x t_y}{t_r^2\sqrt{t_x^2 + t_z^2}}, \\
    \frac{\partial p_y}{\partial t_y} &= +\frac{H}{\pi}\cdot\frac{\sqrt{t_x^2 + t_z^2}}{t_r^2}, \\
    \frac{\partial p_y}{\partial t_z} &= -\frac{H}{\pi}\cdot\frac{t_z t_y}{t_r^2\sqrt{t_x^2 + t_z^2}},
\end{align}
and $\mathbf{W}$ is the rotation part of the $4\times4$ transformation matrix $\mathbf{T}_\text{cw}$ from the world coordinate system to the camera space.
% Let $(\mathbf{A})_{ij}$ denote the element on the $i$-th row and $j$-th column of a matrix $\mathbf{A}$, then $\mathbf{W}$ can be denoted as:
% \begin{equation}\label{equ:W}
%     \mathbf{W} = 
%     \begin{bmatrix}
%         (\mathbf{T}_\text{cw})_{00} & (\mathbf{T}_\text{cw})_{01} & (\mathbf{T}_\text{cw})_{02} \\[8pt]
%         (\mathbf{T}_\text{cw})_{10} & (\mathbf{T}_\text{cw})_{11} & (\mathbf{T}_\text{cw})_{12} \\[8pt]
%         (\mathbf{T}_\text{cw})_{20} & (\mathbf{T}_\text{cw})_{21} & (\mathbf{T}_\text{cw})_{22}
%     \end{bmatrix}
%     .
% \end{equation}
%According to \cite{3dgs:ewasplatting}, we 
The $2\times2$ covariance matrix finally can be obtained by skipping the third row and column of $\Tilde{\mathbf{\Sigma}}$. The forward process could be largely accelerated by the approximation Eq.~\eqref{equ:cov2D}, leading to high-FPS real-time rendering.

Overall, during the tile-based forward rendering process, a whole equirectangular image is partitioned into grids composed of tiles of the same size. First, the center and covariance of 3D Gaussians are projected onto the image screen. Second, each tile counts the 2D Gaussians whose radius of influence covers this tile, generating one instance per influence. Third, all pixels within the same tile are rendered at the same time, each pixel assigned to one thread. These threads cooperatively get the attributes of the Gaussian instances observed by the current tile, then separately accumulate instances for $\alpha$-blending until the pixel has $\alpha=0.9999$. (The stopping threshold is not exactly $1$ for numerical stability considerations.)

\subsection{Backward Optimization}\label{subsec:splatting-backward}
To optimize the world position $\mathbf{m}$, color $c$, rotation $\mathbf{q}$, scale $\mathbf{S}$ and opacity $o$ of 3D Gaussians, we minimize the photometric loss between the rendered image $I_\text{r}$ and ground truth $I_\text{gt}$:
\begin{equation}\label{equ:loss}
    \mathcal{L}(I_\text{r},I_\text{gt}) = (1-\lambda)\left|{ I_\text{r} - I_\text{gt} }\right|_1 + \lambda{(1-\text{SSIM}(I_\text{r}, I_\text{gt}))},
\end{equation}
where $\text{SSIM}(I_\text{r}, I_\text{gt})$ is the structural similarity between two images, and $\lambda$ is a balancing weight factor.

The backward gradient flows from $\mathcal{L}$ to the attributes of 3D Gaussians through the full projection process. To be specific, except for $c$ and $o$ which have nothing to do with the camera model, the gradients of $\mathcal{L}$ over the attributes are what we need to derive and modify for omnidirectional optimization. We can apply the chain rule for multivariable functions to obtain:
\begin{align}
    \frac{\partial \mathcal{L}}{\partial \mathbf{m}} = \sum_{k=1}^M &
    \left[
        \frac{\partial \mathcal{L}}{\partial c}
        \frac{\partial c}{\partial \mathbf{m}}
        +
        \frac{\partial \mathcal{L}}{\partial \alpha_k}
        \frac{\partial \alpha_k}{\partial G_k}
        \left(
            \frac{\partial G_k}{\partial \Tilde{\mathbf{\Sigma}}}
            \frac{\partial \Tilde{\mathbf{\Sigma}}}{\partial \mathbf{J}}
            \frac{\partial \mathbf{J}}{\partial \mathbf{t}}
            \frac{\partial \mathbf{t}}{\partial \mathbf{m}}\right.\right.\notag \\[3pt]
        & \left.\left. +
            \frac{\partial G_k}{\partial \mathbf{p}}
            \frac{\partial \mathbf{p}}{\partial \mathbf{s}}
            \frac{\partial \mathbf{s}}{\partial \mathbf{t}}
            \frac{\partial \mathbf{t}}{\partial \mathbf{m}}
        \right)
    \right] ,\\[3pt]
    \frac{\partial \mathcal{L}}{\partial \mathbf{q}} = \sum_{k=1}^M & \left(
        \frac{\partial \mathcal{L}}{\partial \alpha_k}
        \frac{\partial \alpha_k}{\partial G_k}
        \frac{\partial G_k}{\partial \Tilde{\mathbf{\Sigma}}}
        \frac{\partial \Tilde{\mathbf{\Sigma}}}{\partial \mathbf{\Sigma}}
        \frac{\partial \mathbf{\Sigma}}{\partial \mathbf{R}}
        \frac{\partial \mathbf{R}}{\partial \mathbf{q}}
    \right) ,\\[3pt]
    \frac{\partial \mathcal{L}}{\partial \mathbf{S}} = \sum_{k=1}^M & \left(
        \frac{\partial \mathcal{L}}{\partial \alpha_k}
        \frac{\partial \alpha_k}{\partial G_k}
        \frac{\partial G_k}{\partial \Tilde{\mathbf{\Sigma}}}
        \frac{\partial \Tilde{\mathbf{\Sigma}}}{\partial \mathbf{\Sigma}}
        \frac{\partial \mathbf{\Sigma}}{\partial \mathbf{S}}
    \right) ,
\end{align}
where $M$ is the total number of instances generated by this Gaussian in all tiles. {Note that $\mathbf{R}$ is the rotation matrix converted from quaternion $\mathbf{q}$, which is an inner property determining the Gaussian covariance along with $\mathbf{S}$, and is different from the rotation part $\mathbf{W}$ of transform $\mathbf{T}_\text{cw}$.} Vectors $\mathbf{s}$ and $\mathbf{p}$ are the screen-space and image-space coordinates, respectively. We retain the common portion from gradients given by \cite{3dgs}, and replace the following parts with our omnidirectional gradients:
$\displaystyle{\frac{\partial \Tilde{\mathbf{\Sigma}}}{\partial \mathbf{\Sigma}}},
\displaystyle{\frac{\partial \mathbf{J}}{\partial \mathbf{t}}},
\displaystyle{\frac{\partial \mathbf{p}}{\partial \mathbf{s}}},
\displaystyle{\frac{\partial \mathbf{s}}{\partial \mathbf{t}}}$.
Detailed derivation can be found in our supplementary material.

%------------------------------------------------------------------------
\section{Reconstruction Pipeline}

We illustrate an overview of OmniGS in \cref{fig:pipeline}. Reconstruction starts from a set of SfM-calibrated equirectangular images $\{I_j\}$, each of which has a pose $\mathbf{T}_j$. We obtain initial 3D Gaussians $\mathbb{G}$ from the colored sparse SfM point cloud $\mathbb{P}$. We set their rotation, scale and opacity to unit values as \cite{3dgs} does, then begin a series of optimization iterations. For each iteration, we choose one view $I_j$ from the randomly shuffled $\{I_j\}$, render from $\mathbf{T}_j$ to get $I_\text{r}$, then compute $\mathcal{L}(I_\text{r},I_j)$ and the corresponding backward gradients. After densifying $\mathbb{G}$ based on the following strategies, we advance the optimizer by one step to optimize all 3D Gaussians.

We apply a gradient-based densification control strategy similar to \cite{3dgs}. But instead of using the perspective gradients, we judge whether to densify a Gaussian in sight of its gradient over omnidirectional screen-space location, i.e. $\displaystyle{\frac{\partial \mathcal{L}}{\partial \mathbf{s}}}$, which is derived and recorded during the calculation process in \cref{subsec:splatting-backward}. In detail, for Gaussians with a large enough gradient, if their scales are too large or too small, then they are going to be split or cloned respectively to densify $\mathbb{G}$, enhancing its ability to represent details. We also prune the Gaussians whose scale or screen-space radius is too large, intending to boost the details. Additionally, opacities of all Gaussians also contribute to the densification control. Gaussians with an excessively small opacity are pruned as well. Large opacities are reset to encourage more densification. We execute the densification control process periodically until reaching a certain number of iterations. We conclude the above reconstruction pipeline in \cref{alg:reconstruction}.

\begin{algorithm}[t]
    \caption{Reconstruction Pipeline}
    \label{alg:reconstruction}
    \begin{algorithmic}[1]
            \renewcommand{\algorithmicrequire}{\textbf{Input:}}
            \renewcommand{\algorithmicensure}{\textbf{Output:}}
            \REQUIRE Equirectangular images $\{I_i\}$ with calibrated poses $\{\mathbf{T}_i\}$ and sparse SfM point cloud $\mathbb{P}$
            \ENSURE 3D Gaussians available for novel-view synthesis
            \\ \textit{Initialization} : Create initial 3D Gaussians $\mathbb{G}$ from $\mathbb{P}$
            \FOR {$j = 1$ to maximum iteration}
                \STATE pick a random $i$
                \STATE render $\mathbb{G}$ from $\mathbf{T}_i$ to get $I_\text{r}$ 
                \STATE $\mathcal{L}_j=\mathcal{L}(I_\text{r},I_i)\leftarrow$ Eq.~\eqref{equ:loss}
                \STATE backpropagate $\mathcal{L}_j$
                \IF {$j \leq$ maximum densification iteration}
                    \IF {$j\mod{\text{densification interval}} == 0$}
                        \STATE densify $\mathbb{G}$ by $\displaystyle{\frac{\partial \mathcal{L}_j}{\partial \mathbf{s}}}$
                        \STATE prune $\mathbb{G}$ by $o$
                        \STATE prune $\mathbb{G}$ by $\mathbf{S}$
                    \ENDIF
                    \IF {$j\mod{\text{opacity-resetting interval}} == 0$}
                        \STATE reset large $o$ in $\mathbb{G}$
                    \ENDIF
                \ENDIF
                \STATE advance optimizer by one step
            \ENDFOR
            \RETURN $\mathbb{G}$ 
    \end{algorithmic}
\end{algorithm}

%------------------------------------------------------------------------
\section{Evaluation}
We will report the evaluation results of OmniGS in this section. We compare our reconstruction quality and rendering speed with the baseline SOTA photorealistic 3D reconstruction methods, NeRF \cite{nerf}, Mip-NeRF 360 \cite{nerf:mipnerf360}, 360Roam \cite{nerf:360roam}, Instant-NGP \cite{nerf:ingp}, TensoRF \cite{nerf:tensorf} and EgoNeRF \cite{nerf:egonerf}. We also conducted a cross-validation experiment to confirm the effectiveness of our method compared to the perspective 3DGS.

\begin{figure*}[t]
    \centering
    \includegraphics[width=\linewidth]{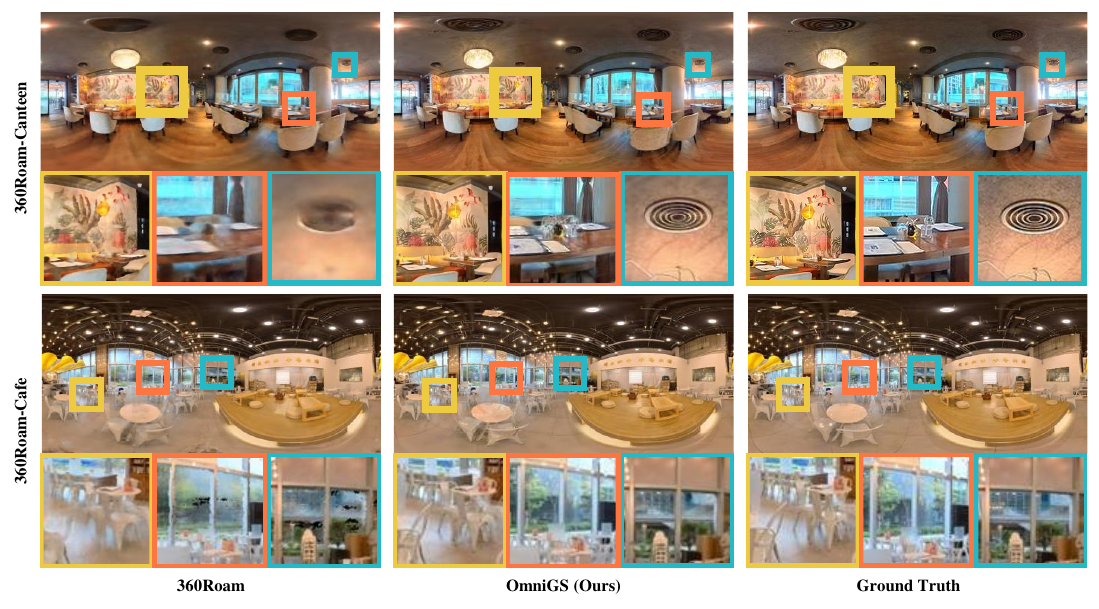}
    \caption{Qualitative comparison example of novel-view synthesis on 360Roam dataset. OmniGS can reconstruct clearer detail structures. It is also free from obvious holes or blurs, as the results in \textit{cafe} show.}
    \label{fig:eval-360roam}
    \vspace{-8pt}
\end{figure*}

\begin{table}[t]
    \centering
    \footnotesize
    \begin{tabular}{c|cccc}
    \hline
    % \textbf{Method} & NeRF\cite{nerf} & Mip-NeRF 360\cite{nerf:mipnerf360} & TensoRF\cite{nerf:tensorf} & Instant-NGP\cite{nerf:ingp} & 360Roam\cite{nerf:360roam} & Ours\\
    \textbf{Method} & {PSNR$\uparrow$} & {SSIM$\uparrow$} & {LPIPS$\downarrow$} & {FPS$\uparrow$} \\
    \hline
    NeRF\cite{nerf} & 22.443 & 0.672 & 0.339 & $<$1 \\
    Mip-NeRF 360\cite{nerf:mipnerf360} & {24.579} & {0.748} & {0.269} & $<$1 \\
    TensoRF\cite{nerf:tensorf} & 15.035 & 0.531 & 0.676 & $<$1 \\
    Instant-NGP\cite{nerf:ingp} & 17.018 & 0.548 & 0.532 & 4 \\
    360Roam\cite{nerf:360roam} & \marksecond{25.061} & \marksecond{0.760} & \marksecond{0.202} & \marksecond{30} \\
    Ours & \markfirst{25.464} & \markfirst{0.806} &  \markfirst{0.141} & \markfirst{121} \\
    \hline
    \end{tabular}
    \caption{Quantitative evaluation results on the 360Roam dataset. We mark the best two results with \colorfirsttext{first} and \colorsecondtext{second}.}\label{tab:eval-360roam}
\vspace{-8pt}
\end{table}

\subsection{Implementation and Experiment Setup}\label{subsec:eval-implementation}

We accomplished OmniGS based on LibTorch framework, which is the C++ version of PyTorch. The tile-based omnidirectional rasterizer was implemented with custom CUDA kernels. As for the usage of datasets, we directly utilized the calibrated camera poses and sparse SfM point clouds contained in the datasets as initial input. But notably, we performed openMVG \cite{sparse:openmvg} SfM on the OmniBlender scenes of EgoNeRF dataset, since these scenes provide no sparse point clouds. For the sake of a fair comparison, we used an RTX-3090 GPU to conduct all experiments on OmniGS, using the provided training and testing split, and gathered the RTX-3090 baseline results reported by the authors of datasets \cite{nerf:360roam} and \cite{nerf:egonerf} unless specifically stated. We evaluated the results in terms of PSNR, SSIM, LPIPS, and rendering FPS {(forward pass inference speed)}, which are common criteria for photorealistic reconstruction. The baselines, except for EgoNeRF, were evaluated in terms of perspective novel-view synthesis, regarding their original objective. Detailed hyperparameter setups are discussed in our supplementary material.

\begin{table*}[t]
    \centering
    \footnotesize
    \tabcolsep=0.15cm
    \begin{tabular}{c|cccc|cccc|cccc}
    \hline
    \multirow{2}{*}{\textbf{Dataset}} & \multicolumn{8}{c|}{\textit{EgoNeRF-OmniBlende}r}  & \multicolumn{4}{c}{\textit{EgoNeRF-Ricoh360}} \\
    & \multicolumn{4}{c|}{Indoor} & \multicolumn{4}{c|}{Outdoor} & \multicolumn{4}{c}{} \\
    \hline
    {\textbf{Method}}
    & {PSNR$\uparrow$} & {SSIM$\uparrow$} & {LPIPS$\downarrow$} & {FPS$\uparrow$} & {PSNR$\uparrow$} & {SSIM$\uparrow$} & {LPIPS$\downarrow$} & {FPS$\uparrow$} & {PSNR$\uparrow$} & {SSIM$\uparrow$} & {LPIPS$\downarrow$} & {FPS$\uparrow$} \\
    \hline
    NeRF\cite{nerf} & 27.660 & 0.756 & 0.425 & $<$1 
                    & 23.630 & 0.686 & 0.458 & $<$1 
                    & 22.780 & 0.663 & 0.538 & $<$1 \\
    Mip-NeRF 360\cite{nerf:mipnerf360} & 27.410 & 0.763 & 0.412 & $<$1
                                       & 25.570 & 0.769 & 0.306 & $<$1
                                       & 24.280 & 0.725 & 0.384 & $<$1\\
    TensoRF\cite{nerf:tensorf} & 29.250 & 0.791 & 0.376 & \marksecond{2}
                               & 25.680 & 0.734 & 0.344 & \marksecond{2}
                               & \marksecond{25.160} & 0.732 & 0.376 & \marksecond{2} \\
    EgoNeRF\cite{nerf:egonerf} & \marksecond{30.230} & \marksecond{0.840} & \marksecond{0.248} & \marksecond{2}
                               & \marksecond{28.810} & \marksecond{0.868} & \marksecond{0.136} & \marksecond{2}
                               & 24.710 & \marksecond{0.746} & \marksecond{0.314} & \marksecond{2}\\
    Ours & \markfirst{35.330} & \markfirst{0.917} & \markfirst{0.072} & \markfirst{115}
         & \markfirst{32.670} & \markfirst{0.919} & \markfirst{0.044} & \markfirst{116}
         & \markfirst{26.032} & \markfirst{0.825} & \markfirst{0.128} & \markfirst{91}\\
    \hline
    \end{tabular}
    \caption{Quantitative evaluation results on the EgoNeRF dataset. We mark the best two results with \colorfirsttext{first} and \colorsecondtext{second}.}
    \label{tab:eval-egonerf}
\end{table*}
\begin{figure*}[t]
    \centering
    \includegraphics[width=\linewidth]{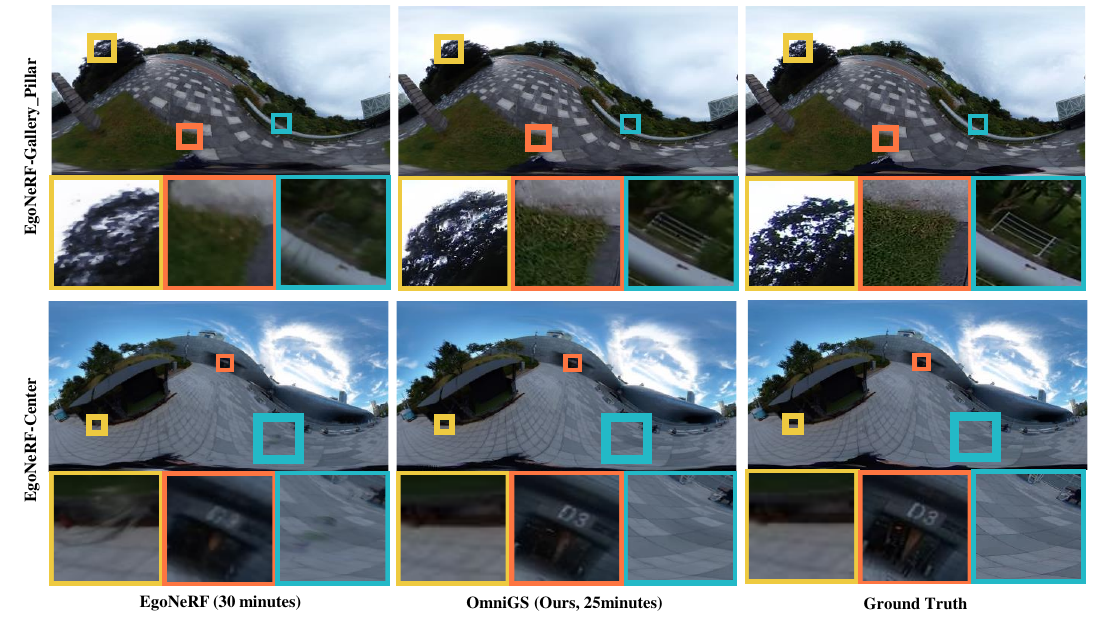}
    \caption{Qualitative comparisons of omnidirectional novel-view synthesis in egocentric scenes. OmniGS can reconstruct the details more sharply and precisely with less training time, i.e. 25 minutes.}
    \label{fig:eval-egonerf}
    \vspace{-8pt}
\end{figure*}

% \subsection{Results and Evaluation}\label{subsec:eval-results}
% \paragraph{On 360Roam dataset \cite{nerf:360roam}}
\subsection{On 360Roam dataset}
360Roam dataset\cite{nerf:360roam} contains 10 complicated real-world indoor scenes captured by a commercial omnidirectional camera fixed on a mobile robot. Note that we skipped the bottom part of the images when computing and backpropagating loss in 360Roam scenes since the base mobile robot is a dynamic object, which is beyond the scope of this paper. So the resolution actually used by omnidirectional evaluation was $712 \times 1520$, for both training and testing.

% overall
We show qualitative comparison examples in \cref{fig:eval-360roam}, and report the quantitative evaluation results in \cref{tab:eval-360roam}. Please refer to our supplementary material to find the per-scene quantitative results. With the help of our derived backward optimization based on explicit geometry representation, OmniGS outperformed the NeRF-based baselines. We gained a slightly higher performance above the state-of-the-art (SOTA) 360Roam which used 2048 multilayer perceptrons and needed several hours to train. We took only less than 25 minutes (32k iterations) to earn similar quality, which was also faster than the other NeRF-based baselines. Our rendering FPS was also at least 4 times higher than the baselines. Note that EgoNeRF needs input images with an egocentric camera motion pattern, and therefore cannot be applied on 360Roam, in which the camera roams randomly in the scenes. In addition, 360-GS \cite{3dgs:360gs} reports an FPS of 60 in indoor scenes with a resolution of $512\times 1024$, which is slower than ours due to its two-stage projection. It also fails to deal with the multiple-room scenes of 360Roam because of its dependence on single-room layout prediction networks.

% \paragraph{On EgoNeRF dataset \cite{nerf:egonerf}}
\subsection{On EgoNeRF dataset}
EgoNeRF\cite{nerf:egonerf} provides 11 OmniBlender simulation scenes and 11 Ricoh360 real-world scenes. The OmniBlender scenes are further classified into 4 indoor and 7 outdoor scenes. We continued to use the provided classification and the original image resolutions, i.e. $1000 \times 2000$ for EgoNeRF-OmniBlender and $960 \times 1920$ for EgoNeRF-Ricoh360, for both training and testing. We trained OmniGS to 32k iterations per scene, which took around 25 minutes. We gathered the results of 10k iterations for EgoNeRF (around 30 minutes), Mip-NeRF 360 (more than 2 hours) and NeRF (more than 5 hours), and 100k iterations for TensoRF (around 40 minutes). Quantitative results are shown in \cref{tab:eval-egonerf}. We also report the per-scene results in our supplementary material. Our method outperformed the SOTA EgoNeRF in terms of both quality and rendering speed (PSNR$+$5.100, SSIM$+$0.077, LPIPS$-$0.176, FPS 57.5 times on OmniBlender-indoor, PSNR$+$3.860, SSIM$+$0.051, LPIPS$-$0.092, FPS 58 times on OmniBlender-outdoor, and PSNR$+$1.322, SSIM$+$0.079, LPIPS$-$0.186, FPS 45.5 times on Ricoh360). We also spent less time on training to acquire such performance. \Cref{fig:eval-egonerf} shows some qualitative comparison examples of omnidirectional rendering, illustrating the ability of OmniGS to reconstruct clearer and sharper details.

\begin{figure}[t]
    \centering
    \includegraphics[width=\linewidth]{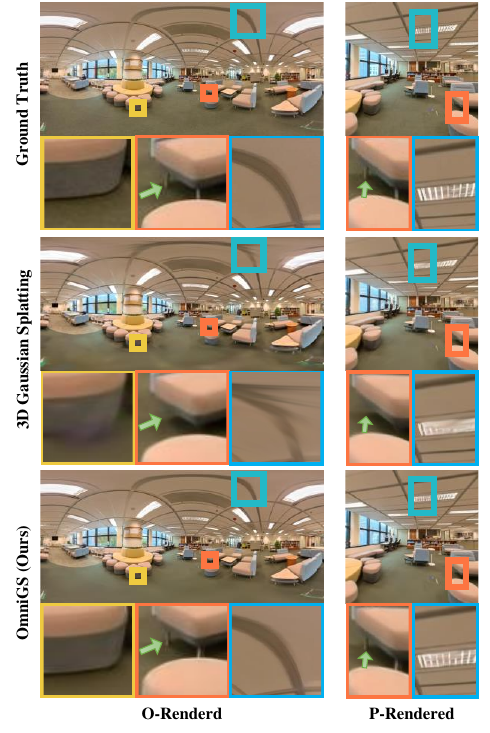}
    \caption{An example result of qualitative evaluation in perspective rendering, on 360Roam-Center. P denotes perspective image, and O denotes omnidirectional image. The perspective 3DGS suffers detail loss and artifacts caused by the limited FoV utilization rate, while OmniGS renders correct omnidirectional novel views that can be cropped to better perspective images.}
    \label{fig:perspective}
    \vspace{-8pt}
\end{figure}

\subsection{Additional Perspective Rendering Evaluation}\label{subsec:eval-perspective}

With a view to validate the effectiveness of our method, we used openMVG to divide each equirectangular image into 6 perspective images and trained the 360Roam scenes again with the perspective 3DGS \cite{3dgs}. We ran the optimization process to the same densification time and total time. Perspectives including the base mobile robot were skipped, for the same reason stated in \cref{subsec:eval-implementation}.

As shown in \cref{tab:perspective}, when considering the tested-as-trained performance, i.e. the P-trained P-tested 3DGS and O-trained O-tested OmniGS, 3DGS was a little better than our OmniGS (P denotes perspective and O denotes omnidirectional). This is because when scene models are evaluated in the form of equirectangular images, the near-pole distortion of real omnidirectional cameras will deteriorate the quantitative results\cite{nerf:egonerf}. The above phenomenon indicates that we should unify the form of testing images before comparing OmniGS with the perspective 3DGS. Taking this into account, we re-evaluated the OmniGS results by cropping the rendered testing views into perspective images, and then compared the two P-tested results. It came out that our method was equipped with the ability to generate better perspective views cropped from the rendered omnidirectional images (PSNR$+$1.812, SSIM$+$0.012, LPIPS$-$0.045). In addition, we rendered the 3DGS-trained models with our omnidirectional rasterizer, then compared the two O-tested results. The omnidirectional novel-view synthesis quality of our models was also higher than that of 3DGS (PSNR$+$2.801, SSIM$+$0.055, LPIPS$-$0.103). Moreover, qualitative results (\cref{fig:perspective}) figure out that the perspective 3DGS tended to lose detail and suffered artifacts due to the low utilization rate of observations. It can use only one limited view at one time. In contrast, when given the same time for training, OmniGS made use of the whole omnidirectional environment in each iteration, reaching more robust densification and faster model convergence. Omnidirectional images rendered from OmniGS-reconstructed models can also be cropped to generate perspective views properly, showing the strong scalability of our method.

%------------------------------------------------------------------------
\section{Conclusion} %and Discussion
In this paper, we present a novel fast photorealistic 3D reconstruction method, named OmniGS, which fully exploits the speed advantage of direct omnidirectional screen-space splatting. We derived the backward gradient and implemented a real-time tile-based omnidirectional rasterizer. Experiment results on various datasets show that OmniGS achieves SOTA reconstruction quality and rendering FPS, even with less training time. 
Compared to the original 3DGS, our method can directly optimize the radiance field using omnidirectional images and realize better perspective view synthesis. We believe OmniGS holds the potential to evolve in various directions, such as integrating OmniGS into omnidirectional visual SLAM systems to perform real-time online photorealistic mapping systems.

\noindent\textbf{Acknowledgements:}
{This work was supported by the China National Key Research and Development Program under Grant 2022YFB3903804.}

\begin{table}
\centering
\footnotesize
\tabcolsep=0.1cm
\begin{tabular}{c|c|c|ccc}
    \hline
    \textbf{Method} & \textbf{Training} &\textbf{Testing}& \textbf{PSNR$\uparrow$} & \textbf{SSIM$\uparrow$} & \textbf{LPIPS$\downarrow$} \\
    \hline
    \multirow{2}{*}{3DGS\cite{3dgs}}  
      & \multirow{2}{*}{P} & \textit{P} & \textit{25.708} & \textit{0.875} & \textit{0.143} \\%\cline{3-6}
      &  & O & 22.663 & 0.751 & 0.244 \\\hline
    \multirow{2}{*}{OmniGS (Ours)} 
      &\multirow{2}{*}{O} & O & \textbf{25.464} & \textbf{0.806} & \textbf{0.141} \\%\cline{3-6}
      & & \textit{P} & \textbf{\textit{27.520}} & \textbf{\textit{0.888}} & \textbf{\textit{0.098}} \\\hline
\end{tabular}
\caption{Quantitative comparison of perspective rendering results on 360Roam Dataset. P denotes perspective image, and O denotes omnidirectional image. {Though OmniGS performs slightly worse when tested as trained, it outperforms 3DGS when the form of testing images are unified into O or \textit{P}, respectively.}}\label{tab:perspective}
\vspace{-8pt}
\end{table}

%%%%%%%%% REFERENCES
{\small
\bibliographystyle{ieee_fullname}
\bibliography{reference}
}

\end{document}